\documentclass{article}
\usepackage{ijcai24}

\usepackage{times}
\usepackage{soul}
\usepackage{url}
\usepackage[hidelinks]{hyperref}
\usepackage[utf8]{inputenc}
\usepackage[small]{caption}
\usepackage{graphicx}
\usepackage{amsmath}
\usepackage{amsthm}
\usepackage{amssymb}
\usepackage{booktabs}
\usepackage{algorithm}
\usepackage{algorithmic}
\usepackage[switch]{lineno}
\usepackage{multirow}

\title{Heterogeneous Causal Metapath Graph Neural Network for Gene-Microbe-Disease Association Prediction}

\author{
Kexin Zhang $^{1}$ 
 \and
Feng Huang $^{1}$ \and
Luotao Liu $^{1}$\and
Zhankun Xiong $^{1}$\and
Hongyu Zhang $^{1,2}$\and \\
Yuan Quan $^{1,2*}$\And
Wen Zhang $^{1,2*}$ \\
\affiliations
$^{1}$ College of Informatics, Huazhong Agricultural University, Wuhan 430070, China\\
$^{2}$ Hubei Key Laboratory of Agricultural Bioinformatics, Huazhong Agricultural University, Wuhan 430070, China\\
\emails
\{kexinzhang, fhuang233, luotaoliu, xiongzk\}@webmail.hzau.edu.cn,\\
\{zhy630, quanyuan, zhangwen\}@mail.hzau.edu.cn
}

\begin{document}

\maketitle

\begin{abstract}

The recent focus on microbes in human medicine highlights their potential role in the genetic framework of diseases. To decode the complex interactions among genes, microbes, and diseases, computational predictions of gene-microbe-disease (GMD) associations are crucial. Existing methods primarily address gene-disease and microbe-disease associations, but the more intricate triple-wise GMD associations remain less explored. In this paper, we propose a \textbf{H}eterogeneous \textbf{C}ausal \textbf{M}etapath \textbf{G}raph \textbf{N}eural \textbf{N}etwork (HCMGNN) to predict GMD associations. HCMGNN constructs a heterogeneous graph linking genes, microbes, and diseases through their pairwise associations, and utilizes six predefined causal metapaths to extract directed causal subgraphs, which facilitate the multi-view analysis of causal relations among three entity types. Within each subgraph, we employ a causal semantic sharing message passing network for node representation learning, coupled with an attentive fusion method to integrate these representations for predicting GMD associations. Our extensive experiments show that HCMGNN effectively predicts GMD associations and addresses association sparsity issue by enhancing the graph's semantics and structure.

\end{abstract}

\section{Introduction}
Contemporary human genetics focuses on unraveling the links between genotypes and diseases \cite{lappalainen2021variant}. Recent research increasingly connects genetic variant-related diseases with human microbes \cite{hall2017human,manor2020health,QUAN2023451}. Furthermore, there's growing evidence of causal relations among genes, microbes, and diseases, offering insights into complex disease pathogenesis from a multifactorial angle \cite{luca2018functional}. Classical bioinformatics approaches, like multi-omics integration, while used for gene-microbe-disease (GMD) association discovery \cite{priya2022identification,markowitz2022microbiome}, lack efficiency at scale. Machine learning-based methods, proposed as an efficient alternative, typically focus only on pairwise associations, such as gene-disease or microbe-disease associations \cite{ata2021recent,wen2021survey}. Their limitation is not considering the third entity's role in these associations. In contrast, exploring triple-wise GMD associations provides a more comprehensive understanding of biological mechanisms.

Several approaches are applicable for triple-wise association prediction, including tensor decomposition \cite{wang2018drug,chen2019modeling,chen2020learning}, hypergraph neural networks \cite{cheng2022ihgnn,liu2022multi}, and n-ary relational knowledge graph embedding methods \cite{wen2016representation,fatemi2019knowledge,liu2021role}. Models based on tensor decomposition, like those proposed by \cite{wang2018drug} and \cite{chen2020learning}, have been used for predicting drug-protein-disease associations. \cite{liu2022multi} introduced a method for anti-cancer drug synergy prediction using a multi-way relation-enhanced hypergraph, while \cite{cheng2022ihgnn} developed an interactive hypergraph neural network for personalized product search. \cite{wen2016representation} focused on representation learning in n-ary relational knowledge graphs. However, these methods typically treat triple-wise associations as instances of three entities without adequately considering the directional relations that indicate causal interactions among them.

With the advancement of microbiome genome-wide association studies, bidirectional interactions between genes and microbes have garnered attention \cite{nichols2021relationship}. Furthermore, causal relations between microbes and diseases have also been validated through causal inference methods such as Mendelian Randomization \cite{lv2021causal}. Consequently, there exist intricate causal modes among genes, microbes, and diseases \cite{hughes2020genome,xu2020interplay}, as summarized by \cite{luca2018functional}. Illustratively, {\it NOD2} gene heightens the susceptibility to inflammatory bowel disease through the modulation of {\it Enterobacteriaceae} abundance \cite{hall2017human}. Despite the insightful triple-wise associations revealed by these causal modes, modeling such interactions presents inherent challenges. Heterogeneous graphs, key in modeling complex systems, use metapaths to reveal specific connections and hidden patterns among entities \cite{shi2018heterogeneous}. While traditional metapath-based methods on these graphs typically model relations between two nodes, how to extend these methods to encompass triple-wise relations is underexplored. Such an extension has significant potential in GMD associations prediction, as the intricate causal modes among genes, microbes, and diseases can be comprehensively mapped using metapaths, termed causal metapaths, which can be used for learning interactive causal semantic information (more details of the causal metapaths description are presented in the Appendix Section 1).

In our study, we propose a \textbf{H}eterogeneous \textbf{C}ausal \textbf{M}etapath \textbf{G}raph \textbf{N}eural \textbf{N}etwork (HCMGNN) to predict GMD associations. HCMGNN constructs a heterogeneous graph that connects genes, microbes, and diseases through their pairwise associations, and utilizes six specific causal metapaths to extract directed causal subgraphs from the heterogeneous graph, which facilitate the multi-view analysis of causal relations among three types of entities (genes, microbes, and diseases). Within each subgraph, we apply a causal semantic sharing message passing network for learning node representations, further augmented by an attentive fusion method for integrating these representations in predicting GMD associations. Extensive experiments demonstrate that HCMGNN achieves satisfactory results. In summary, the main contributions of this work are described as follows:
\begin{itemize}
    \item To our knowledge, HCMGNN is the first metapath-based heterogeneous graph framework for triple-wise association prediction, which can efficiently capture structural and semantic information for large-scale discovery of potential associations among genes, microbes, and diseases. 
    \item To model the triple-wise relations among genes, microbes, and diseases, we consider six distinct metapaths to represent their interaction modes, which have been observed in the real world. These metapaths serve as a foundation for generating various subgraphs, enabling us to capture a rich array of semantic information.
    \item Within each subgraph, we employ a causal semantic sharing message  passing network for node representation learning, coupled with an attentive fusion method to integrate multi-view node representations for predicting GMD associations.
\end{itemize}

\section{Related Work} 
\subsection{Biomedical Association Mining}
Within the biomedical domain, research on disease-related association mining primarily highlights the functional information of genes \cite{ata2021recent}. Recently, several studies have shifted their focus towards elucidating the role of microbes in complex diseases, thereby discovering novel associations between microbes and diseases \cite{wen2021survey}. Consequently, the exploration of associations between diseases and multiple entity types has emerged as a burgeoning field in biomedicine. Here, we summarize the triple-wise association prediction methods in bioinformatics in recent years.
Examples include tensor decomposition models, applied by \cite{wang2018drug} and \cite{chen2020learning} for predicting drug-protein-disease associations; Hypergraph neural networks, employed by \cite{liu2022multi} to predict anti-cancer drug synergy; N-ary relational knowledge graph embedding methods, utilized by \cite{wen2016representation} to learn representation for predicting n-ary facts. Compared to the above methods, we model the triple-wise associations based on heterogeneous graph framework, and capture higher-order interactions by causal modes among different types of entities.

\subsection{Metapath-based HGNNs} 
Heterogeneous graphs, characterized by diverse node and edge types, exhibit pivotal advancement in modeling complex systems. Recently, heterogeneous graph neural networks (HGNNs) have attracted extensive attention from different research fields due to their great potential in learning and generalizing patterns within heterogeneous structures and have been applied to various downstream tasks, such as node classification \cite{HAN,wang2021self}, recommender systems \cite{yan2021attention,yang2021hyper} and bioinformatics \cite{li2022hierarchical,zhang2023mhtan,li2023metapath}. 
To distill diverse semantic information within a specific context from heterogeneous graph, most existing HGNNs commonly employ metapaths, which refers to a sequence of relations connecting different types of nodes in heterogeneous graph.

Metapath-based HGNNs can be broadly divided into two categories according to their metapath-induced message passing formation: (1) aggregating metapath-induced neighbors \cite{HAN,HGBER,wang2021self,yan2023osgnn}. Usually, the target nodes are head entities in metapath instances and their metapath-induced neighbors are tail entities. For instance, HAN generates homogeneous graphs based on several symmetric metapaths, where graph attention layers are adopted to update node embeddings \cite{HAN}.
(2) aggregating metapath instances \cite{MAGNN,zheng2022explainable}. For example, MAGNN update target nodes by aggregating all symmetric metapath instances with consistent head and tail node types involving the target node \cite{MAGNN}. Almost all abovementioned traditional metapath-based HGNNs predominantly concentrate on capturing high-order relations within homogeneous node pairs, while neglecting intricate interactions among entities of distinct types. 

\section{Methodology}  
\subsection{Problem Formulation} 
Let $\mathcal{N}$, $\mathcal{M}$ and $\mathcal{D}$ denote the set of genes, microbes, and diseases, respectively. Based on the verified pairwise associations among them, i.e. gene-microbe, gene-disease and microbe-disease associations, we can construct a GMD heterogeneous graph $\mathcal{G = \{V, E\}}$, in which $\mathcal{V = N\cup M \cup D}$ is the node set and $\mathcal{E}$ is a set of edges representing the three types of pairwise associations. Each node $v \in \mathcal{V}$ is characterised by a feature vector denoted as $\boldsymbol{x}_v$. 
All feature vectors of gene nodes, microbe nodes and disease nodes are compiled into $\boldsymbol{X}_\mathcal{N} \in \mathbb{R}^{|\mathcal{N}| \times F_{\mathcal{N}}}$, $\boldsymbol{X}_\mathcal{M} \in \mathbb{R}^{|\mathcal{M}| \times F_\mathcal{M}}$ and $\boldsymbol{X}_\mathcal{D} \in \mathbb{R}^{|\mathcal{D}| \times F_\mathcal{D}}$. 
We consider a triplet $(n,m,d) \in \mathcal{N} \times \mathcal{M} \times \mathcal{D}$ has a triple-wise association only when all three entities have associations with each other. 
Such a triplet is assigned a label $y=1$ representing the existence of the triple-wise association.
Other triplets are labeled as $y=0$, which means potential associations undiscovered before. 
Our goal is to build an effective in-silico model to prioritizes potential triple-wise associations based on the GMD heterogeneous graph.
\subsection{Model Architecture}
The overall framework of the HCMGNN model is shown in Figure 1. HCMGNN mainly includes: (1) feature transformation; (2) causal subgraph generation; (3) intra-subgraph message passing; (4) inter-subgraph attentive fusion; (5) model training.
\begin{figure*}[t]
\centering
\includegraphics[width=2.03\columnwidth]{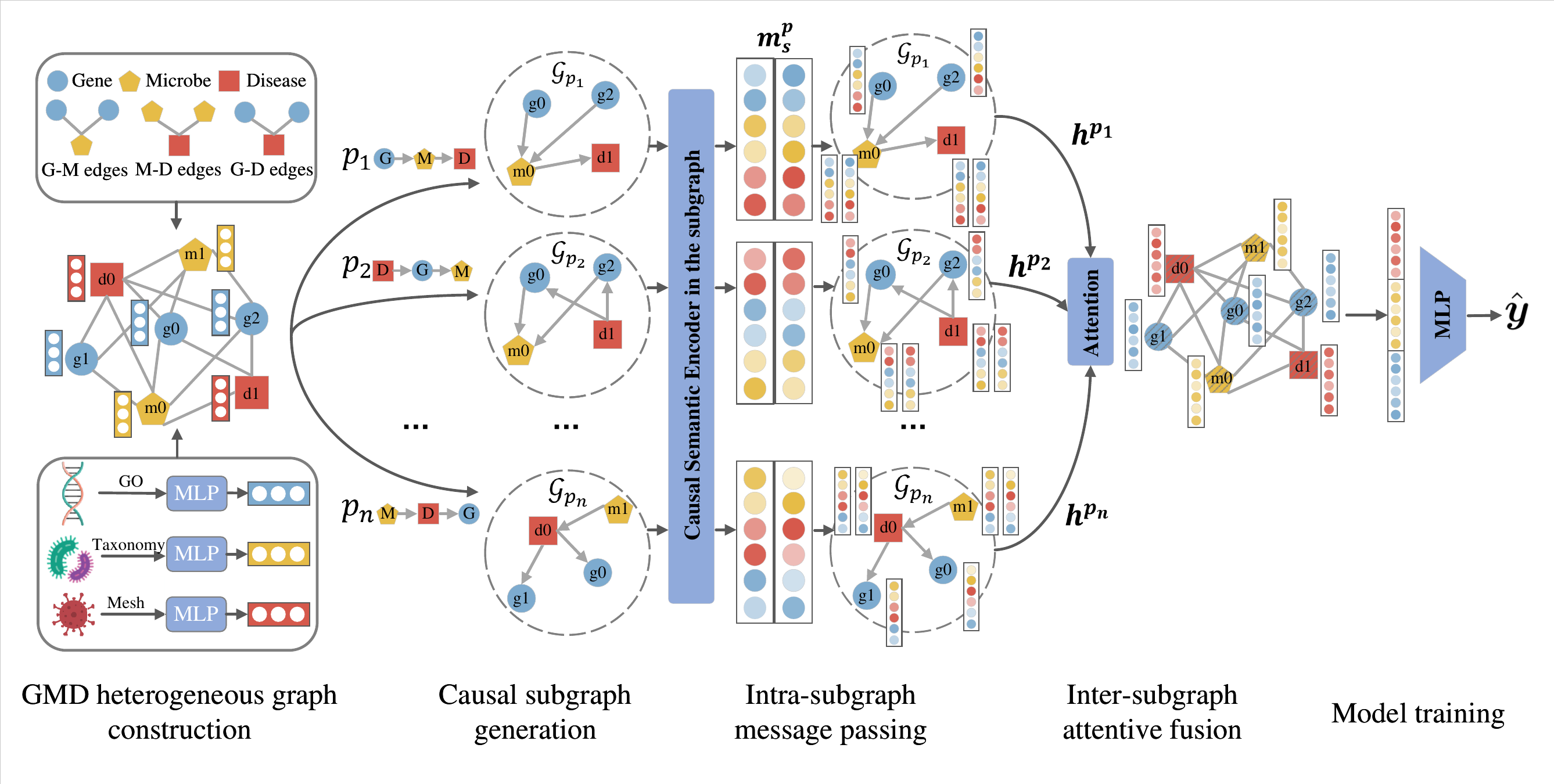} 
\caption{Overall architecture of HCMGNN. HCMGNN mainly includes:
(1) feature transformation; (2) causal subgraph generation; (3) intra-subgraph message passing; (4) inter-subgraph attentive fusion; (5) model training.}
\label{fig4}
\end{figure*}
\subsection{Feature Transformation}
In the GMD heterogeneous graph, different node types represent distinct biological entity categories. Different types of node feature vectors are compiled by diverse means of data collection and preprocessing, and in result they have unequal dimension and scale. Therefore, we simply operate a type-specific linear transformation on each type of node features to map all nodes into the same latent space. Let $\mathcal{K}$ be an option from $\mathcal{M}$, $\mathcal{N}$ and $\mathcal{D}$. The type-specific linear transformation is formulated by: 

\begin{equation}
    \boldsymbol{h}_v=W_\mathcal{K} \cdot \boldsymbol{x}_v
\end{equation}
where $W_\mathcal{K} \in \mathbb{R}^{F^{\prime} \times F_\mathcal{K}}$ is the learnable weight in the linear transformation for genes, microbes or diseases, $\boldsymbol{x}_v \in \mathbb{R}^{F_\mathcal{K}}$ is node feature vector of node $v\in \mathcal{K}$, and $\boldsymbol{h}_v \in \mathbb{R}^{F^{\prime}}$is the projected feature vector (supplementary details about node features are presented in the Appendix Section 2). 

\subsection{Causal Subgraph Generation}
As mentioned above, the causal modes among genes, microbes, and diseases can be effectively represented using metapaths. To ensure that each metapath conveys the interaction direction among the three entity types, we prefine six metapaths with causal modes, named causal metapaths, encompassing {\it G-M-D}, {\it G-D-M}, {\it D-M-G}, {\it D-G-M}, {\it M-D-G}, and {\it M-G-D}. For example, {\it G-M-D} indicates that genes have the potential to influence complex diseases through the modulation of microbial changes \cite{luca2018functional} (more details and underlying implications are presented in Appendix Section 1). Next, we extract six directed subgraphs from the heterogeneous graph $\mathcal{G}$ under the guidance of metapaths to enable the model to learn informative causal context. 

Specifically, we consider all edges in $\mathcal{E}$ are bi-directional such that $\mathcal{E}$ contains six types of directed edges corresponding to six kinds of relations. Note that each causal metapath is composed of two relations. Let $\mathcal{E}_r \subset \mathcal{E}$ is the set of edges belonging to relation $r$. Given a metapath $p$ that consists of two relations $r$ and $r^\prime$, we can generate a directed subgraph $\mathcal{G}_p=\{\mathcal{V}, \mathcal{E}_p\}$ of $\mathcal{G}$, where $\mathcal{E}_p=\{\mathcal{E}_{r}, \mathcal{E}_{r^\prime}\}$. Then, the causal subgraph comprises all metapath instances of $p$ and each instance implies causal semantic information that the head node exacts impact on the tail node through the intermediate node. 

\subsection{Intra-Subgraph Message Passing}

In this section, we present a message passing network with causal semantic sharing mechanism to update node embeddings in each subgraph. The key idea is to share the causal semantic information in each metapath instance across its head, intermediate and tail nodes. Formally, we define that $\mathcal{S}_p$ is the set of instances of metapath $p$, and $\mathcal{S}_p(v) \subset \mathcal{S}_p$ is the set of instances involving node $v$. Then, the node embedding $\boldsymbol{h}^p_v$ of $v$ in the view of subgraph $\mathcal{G}_p$ is given by:
\begin{equation}
    \begin{aligned}   
    \boldsymbol{m}^p_{s}&=\Phi(\{\boldsymbol{h}_u\mid u\in \mathcal{V}^p_s\}) \\
        \boldsymbol{h}^p_v &= \Psi(\boldsymbol{h}_v, \{\boldsymbol{m}^p_{s}\mid s\in\mathcal{S}_p(v)\})
    \end{aligned} 
\end{equation}
where $\boldsymbol{m}^p_{s}$ is the embedding of the metapath instance $s\in\mathcal{S}_p$ and $\mathcal{V}^p_s$ contains three nodes in $s\in\mathcal{S}_p$. Obviously, the embedding of metapath instance $s$ serves as the message shared with every node of it. What follows, we introduce detailed implementation for the causal semantic encoder $\Phi(\cdot)$ and the message aggregation function $\Psi(\cdot)$ in our work.

\subsubsection{Causal Semantic Encoder}
Recall that all metapath instances contain three nodes respectively from $\mathcal{N}$, $\mathcal{M}$ and $\mathcal{D}$, connected by two kinds of relations. To embed causal semantic information of the metapath instance $s$ into $\boldsymbol{m}_{s}$, $\Phi(\cdot)$ requires not only the fusion of node embeddings but also capability in capturing relational patterns. We adopt a approach proposed by \cite{sun2018rotate,MAGNN} to model the causal relations. Suppose that a metapath instance $s\in \mathcal{S}_p$ is defined by $h \stackrel{r_{he}}{\longrightarrow} e \stackrel{r_{et}}{\longrightarrow} t$, where $h$, $e$ and $t$ are the head, intermediate and tail node; and $r_{he}$ and $r_{et}$ are the relations. Then, $\boldsymbol{m}^p_{s}$ is obtained by: 

\begin{equation}
    \begin{aligned}
\boldsymbol{m}^p_{s}=\frac{1}{3}((\boldsymbol{h}_h\odot\boldsymbol{r}_{he}+\boldsymbol{h}_e)\odot\boldsymbol{r}_{et}+\boldsymbol{h}_t)
\end{aligned}
\end{equation}
where $\boldsymbol{r}_{he}$ and $\boldsymbol{r}_{et}$ are both learnable parameters for embedding relations, and $\odot$ is Hadamard product.
\subsubsection{Message Aggregation}

Considering different metapath instances will contribute to the representation of the node in varying extents, here we utilize the graph attention layer \cite{GAT} to aggregate causal semantics by taking all the metapath instances involving node $v$ as its neighbors. Specifically, we have:

\begin{equation} 
        \boldsymbol{h}^p_v=\sigma\left(\sum_{s\in \mathcal{S}_p(v)} \alpha^p_{sv} \cdot \boldsymbol{m}^p_s\right)
\end{equation}
where $\sigma$ is an activation function and $\alpha^p_{sv}$ is the attention value indicating the contribution of semantic message $\boldsymbol{m}^p_s$ from metapath instance $s\in \mathcal{S}_p(v)$ to node $v$ in the subgraph $\mathcal{G}_p$. The attention value $\alpha^p_{sv}$ for $s\in \mathcal{S}_p(v)$ can be calculated by:
\begin{equation} 
        \alpha^p_{sv}=\frac{\exp \left(\operatorname{LeakyReLU}\left(\boldsymbol{a}_p^{\!\scriptscriptstyle\top} \cdot\left[\boldsymbol{h}_v \| \boldsymbol{m}^p_s\right]\right)\right)}{\sum_{s^\prime\in \mathcal{S}_p(v)} \exp \left(\operatorname{LeakyReLU}\left(\boldsymbol{a}_p^{\!\scriptscriptstyle\top} \cdot\left[\boldsymbol{h}_v \| \boldsymbol{m}^p_{s^\prime}\right]\right)\right)}
\end{equation}
where $\boldsymbol{a}_{ p}^{\!\scriptscriptstyle\top}$ is the trainable parameterized attention vector for metapath $p$, and $\|$ represents concatenation operation of vector. To stabilize the learning process and reduce the high variance of the attention coefficient caused by graph heterogeneity, we employ multi-head attention similarly to \cite{GAT} that concatenates multi-view node embeddings from multiple attention heads.

\subsection{Inter-Subgraph Attentive Fusion}
Now that we obtain multiple views of node embeddings from six subgraphs generated before, a type-specific attentive fusion operator is exploited to merge the multi-view information. Concretely, suppose that $\mathcal{P}$ is the set of our predefined metapaths and $p\in \mathcal{P}$, the attention coefficient $e^{p}_{\scriptscriptstyle\mathcal{K}}$ assigned for each node in $\mathcal{K}$ in the subgraph $\mathcal{G}_p$ is computed by:
\begin{equation}
    e^{p}_{\scriptscriptstyle\mathcal{K}}=\frac{1}{\left|\mathcal{K}\right|} \sum_{v \in \mathcal{K}} \boldsymbol{q}_{\scriptscriptstyle\mathcal{K}}^{\!\scriptscriptstyle\top} \cdot \tanh \left(\boldsymbol{W}^{\prime}_{\mathcal{K}} \cdot \boldsymbol{h}_v^{p}+\boldsymbol{b}_{\scriptscriptstyle\mathcal{K}}\right)
\end{equation}
where $\boldsymbol{q}_{\scriptscriptstyle\mathcal{K}}^{\!\scriptscriptstyle\top}$ is a parameterized query vector for node type $\mathcal{K}$, and $\boldsymbol{W}^{\prime}_{\mathcal{K}}$ and $\boldsymbol{b}_{\scriptscriptstyle\mathcal{K}}$ are the learnable weight matrix and bias vector respectively. 

Note that the attention coefficient is shared across all nodes in $\mathcal{K}$, which measures an equal contribution of causal semantic information behind $p$ to all node embeddings of node type $\mathcal{K}$. Then we further normalize these attention coefficients with a softmax layer.
\begin{equation}
    \beta^{p}_{\scriptscriptstyle\mathcal{K}} =\frac{\exp \left(e^{p}_{\scriptscriptstyle\mathcal{K}}\right)}{\sum_{p^{\prime}\in \mathcal{P}} \exp \left(e^{p^{\prime}}_{\scriptscriptstyle\mathcal{K}}\right)}
\end{equation}

Finally, we weightedly sum all views of embeddings of node $v \in \mathcal{K}$ using the weights given by the normalized attentions to obtain the final node embedding $\boldsymbol{z}_v$.
\begin{equation}
    \boldsymbol{z}_v =\sum_{p\in \mathcal{P}} \beta^{p}_{\scriptscriptstyle\mathcal{K}} \cdot \boldsymbol{h}^p_v
\end{equation}

\begin{table*}[htbp]
\centering
\setlength{\tabcolsep}{1.6mm}
\begin{tabular}{ccccccccc}%
\toprule          
Type & Method & Hit@1 & Hit@3 & Hit@5 & NDCG@1 & NDCG@3 & NDCG@5 & MRR \\
\midrule
& RF & 0.4880 & 0.7996 & 0.8964 & 0.4880 & 0.6713 & 0.7115 & 0.6614 \\
Machine Learning & MLP & 0.5321 & 0.8277 & 0.9122 & 0.5321 & 0.7060 & 0.7409 & 0.6937 \\
& XGBoost & 0.5719 & 0.8339 & 0.9132 & 0.5719 & 0.7269 & 0.7598 & 0.7184 \\
\hline 
\rule{0pt}{10pt} 
& Tucker & 0.7500 & 0.8021 & 0.8271 & 0.7500 & 0.7805 & 0.7907 & 0.7890 \\
Tensor Decomposition & CP & 0.7914 & 0.8403 & 0.8578 & 0.7914 & 0.8208 & 0.8278 & 0.8271 \\
& NeurTN & 0.5424 & 0.8391 & 0.9168 & 0.5424 & 0.7169 & 0.7493 & 0.7017 \\
\hline 
\rule{0pt}{10pt} 
Knowledge Graph & mTransH & 0.6943 & 0.8326 & 0.8779 & 0.6943 & 0.7757 & 0.7944 & 0.7771 \\
Embedding & RAM & 0.7019 & 0.8357 & 0.8844 & 0.7019 & 0.7806 & 0.8007 & 0.7830 \\
\hline
\rule{0pt}{10pt} 
& HAN & 0.6771 & 0.8731 & 0.9122 & 0.6771 & 0.7934 & 0.8096 & 0.7819 \\
Graph & MAGNN & 0.7432 & 0.8863 & 0.9362 & 0.7432 & 0.8262 & 0.8486 & 0.8256 \\
Neural Network & HypergraphSynergy & 0.7853 & 0.9326 & 0.9563 & 0.7853 & 0.8773 & 0.8863 & 0.8656 \\
& HCMGNN & \textbf{0.7947} & \textbf{0.9417} & \textbf{0.9641} & \textbf{0.7947} & \textbf{0.8827} & \textbf{0.8923} & \textbf{0.8712} \\
\bottomrule
\end{tabular}
\caption{Performances of HCMGNN and baselines evaluated by Hit@n, NDCG@n and MRR in 5-fold CV.}
\label{table1}
\end{table*}

\subsection{Model Training}
Through the modules introduced above, we obtain highly-abstracted node representations for genes, microbes, and diseases, denoted as $\boldsymbol{z}_n$ for node $n\in \mathcal{N}$, respectively $\boldsymbol{z}_m$ for node $m\in \mathcal{M}$ and $\boldsymbol{z}_d$ for node $d\in \mathcal{D}$. After that, we feed the concatenation of $\boldsymbol{z}_n$, $\boldsymbol{z}_m$ and $\boldsymbol{z}_d$ into a MLP layer followed by a Sigmoid layer for calculating the probability of the existence of GMD association among the triplet $(n,m,d)$.

\begin{equation}
\hat{y}=\operatorname{Sigmoid}\left(\operatorname{MLP}\left(\boldsymbol{z}_n\left\|\boldsymbol{z}_m\right\| \boldsymbol{z}_d\right)\right)
\end{equation}

Then, we adopt the loss function used in \cite{wang2022heterogeneous} to optimize our model, which is formulated as:
\begin{equation}
L=(1-\gamma)\|\boldsymbol{y} \odot(\boldsymbol{y}-\hat{\boldsymbol{y}})\|^2+\gamma\|(1-\boldsymbol{y}) \odot(\boldsymbol{y}-\hat{\boldsymbol{y}})\|^2
\end{equation}
where $\gamma$ is a balance coefficient, ${\| \cdot \|}^2$ is the $L_2$ norm, $\boldsymbol{\hat{y}}$ is a vector storing prediction scores for all triplets in training set and $\boldsymbol{y}$ represents the corresponding ground truth labels.

\section{Results}
\subsection{Experimental Settings}
\subsubsection{Dataset}
We construct a dataset of gene-microbe-disease associations using several public databases. The gene-microbe associations are collected from GIMICA \cite{tang2021gimica} and gutMGene \cite{cheng2022gutmgene}. The microbe-disease associations are derived from MicroPhenoDB \cite{yao2020microphenodb}, GutMDdisorder \cite{cheng2020gutmdisorder} and Peryton \cite{skoufos2021peryton}. And the gene-disease associations are obtained from the DisGeNET \cite{pinero2016disgenet}. Then we merge the gene-microbe associations, microbe-disease associations, and gene-disease associations
into triplets $<$gene-microbe-disease$>$. After filtering, we obtain a total of 3431 triplets, involving 301 genes, 176 microbes, and 153 diseases. 

\subsubsection{Baselines}
In order to verify the effectiveness of our model, we compare HCMGNN with baselines as follows (more details of baselines can be found in Appendix Section 3):

\begin{itemize}
    \item Random Forest (RF), multilayer perception (MLP) and XGBoost are classic machine learning methods that utilize concatenated features of entities for prediction.
    \item Tucker and CP are tensor decomposition models, which utilize multi-dimensional tensors to represent and reconstruct the high-order relations \cite{kolda2009tensor}.
    \item NeurTN is a neural tensor model combining tensor algebra and deep neural network that captures associations among drugs, targets, and diseases \cite{chen2020learning}.
    \item mTransH and RAM are n-ary relational knowledge graph embedding methods. mTransH is a  translational distance model to handle multiple relation types \cite{wen2016representation}. RAM introduce a role-aware modeling to learn the semantic relatedness among roles for completing n-ary facts in knowledge graph \cite{liu2021role}.
    \item HAN and MAGNN are metapath-based heterogeneous graph neural network models for node classification and link prediction \cite{HAN,MAGNN}. 
    \item HypergraphSynergy is a multi-way relation-enhanced hypergraph representation learning method for predicting drug-drug-cell line associations \cite{liu2022multi}.

\end{itemize}

\subsubsection{Implementation Details}
We randomly split the dataset into 90\% cross-validation (CV) set and 10\% independent test set. Then we perform the 5-fold CV on the CV set to train the model and optimize the hyper-parameters, and evaluate model performance on the independent test set. In each CV, we randomly select four out of five positive samples in the CV set and generate an equal number of negative samples as the training set. When verifying and testing the model, we generate 30 corresponding negative samples for each test triplet, so as to sort the prediction scores of all samples. To evaluate the performance of the model, we focus on three evaluation metrics commonly used in recommender systems and knowledge graph embeddings model \cite{chen2020learning}: hit ratio (Hit@n) and normalized discounted cumulative gain (NDCG@n) with n=1, 3, 5 and mean reciprocal rank (MRR). For training our model, we set the balance coefficient of the loss function to 0.7, employ the Adam optimizer with a learning rate set to 0.005 to optimize the model, and adopt early stopping mechanism with a patience of 50 to terminate the training early (the hyper-parameter sensitivity analysis are provided in Appendix Section 4). Source code and dataset of HCMGNN are available online at \url{https://github.com/zkxinxin/HCMGNN}.
\subsection{Performance Comparison} 

In this section, We compare HCMGNN with baseline methods in 5-fold CV and independent test experiments. As shown in Table 1, HCMGNN outperforms other baselines under all evaluation metrics on the CV set, demonstrating its effectiveness in GMD association prediction. We also have the following observations: (1) Compared with RF, MLP, and XGBoost, which only consider entities feature information, HCMGNN achieves a significant performance improvement, which indicates that the rich structural and semantic information learned by heterogeneous graph enhances the capability of node representation.
(2) Compared with Tucker, CP, and NeurTN, which model higher-order associations through multi-dimensional tensors, HCMGNN considers the relational directionality modeled by causal metapaths, making it more conducive to comprehensively learning triple-wise interaction information than tensors. 
(3) mTransH and RAM, which model n-ary relational knowledge graphs by translation distance and role semantics, respectively, surpass the above methods on most metrics. HCMGNN achieves further improvement based on them, which may be due to that HCMGNN jointly considers the topological neighborhood structure of heterogeneous graph and triple-wise interaction information, which is beneficial to aggregate more comprehensive information for nodes.
(4) MAGNN, which learn node embeddings by aggregating metapath instances in hetergeneous graph, performs superior to HAN. However, HCMGNN still make improvement over them, which implies that compared with the traditional metapath-based HGNNs, HCMGNN is better able to model the interaction of multi-view triple-wise interactions by utilizing causal metapaths and update node representations in a shared semantics way comprehensively.
(5) HypergraphSynergy, which utilizes hypergraphs to represent triple-wise associations and exhibits better performance than other baselines. However, HCMGNN still outperforms it across all metrics, suggesting that metapaths, as an impactful means to connect high-order relations, can effectively learn representations of nodes from multiple causal views. These results provide valuable implications for addressing the triple-wise association problems.

In addition, according to the performance of HCMGNN and baselines in independent test set (experiment result can be found in Appendix Section 5), we find that the general trend of the results is consistent with the CV set, HCMGNN outperforms all baselines under every evaluation metrics and exceeds a state-of-the-art method HypergraphSynergy by 2.0\%, 1.8\%, and 1.6\% in Hit@1, Hit@3 and MRR, which verifies its superiority in predicting the GMD triple-wise association.

\begin{figure}[t]
\centering
\includegraphics[width=1\columnwidth]{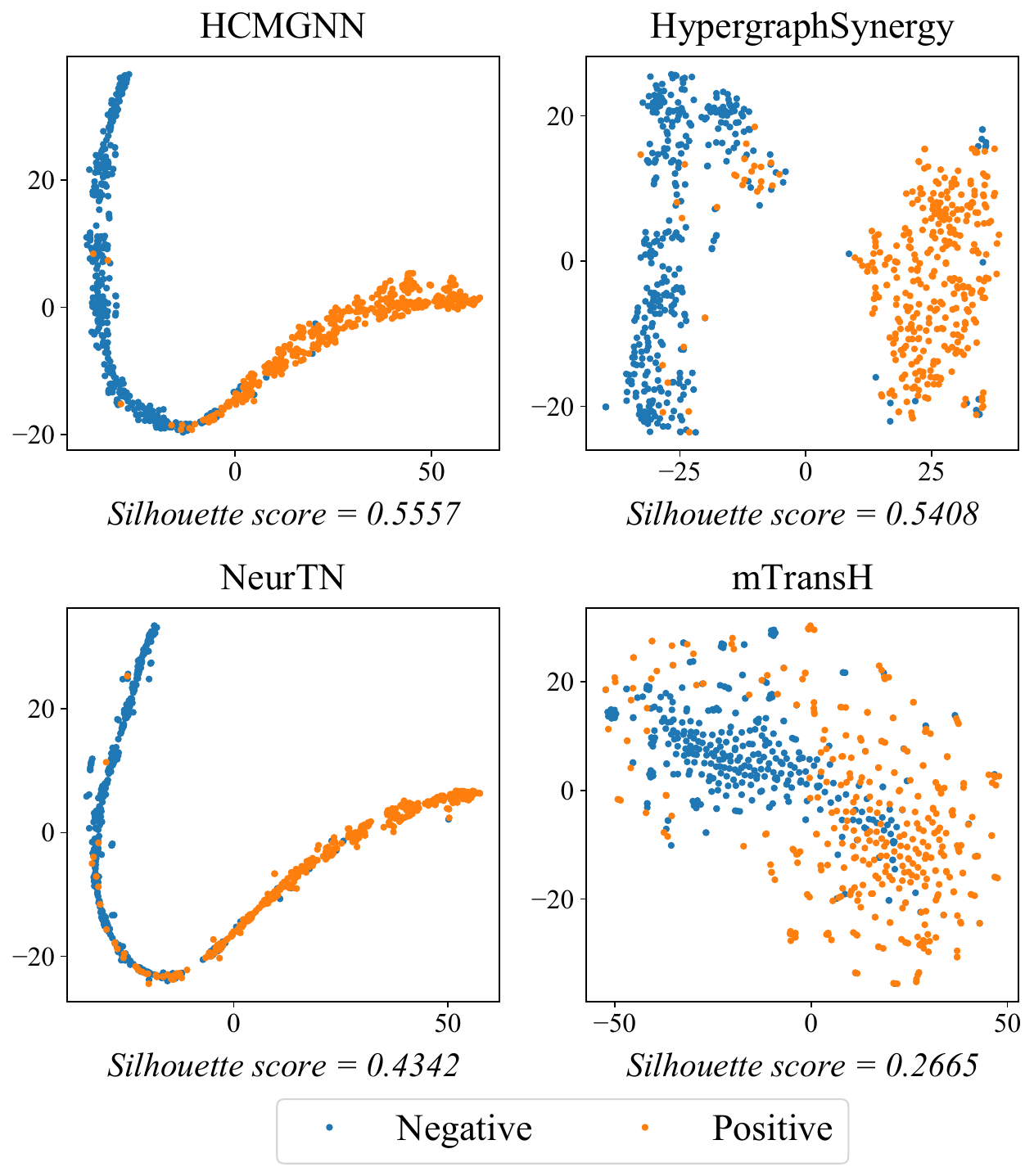} 
\caption{The t-SNE visualization of four models.}
\label{fig1}
\end{figure} 

To intuitively verify the powerful ability of HCMGNN to learn GMD representation, we performed t-SNE visualization of triplets embeddings learned by HCMGNN and several state-of-the-art baselines from independent test set. Figure 2 shows the embedding visualization results of positive and negative samples of HCMGNN, hypergraph neural network (HypergraphSynergy) and tensor decomposition with deep neural networks (NeurTN) and n-ary relational knowledge graph embedding model (mTransH). It can be seen that HCMGNN can better distinguish positive samples from negative samples compared with the comparison methods and achieves the best silhouette score of 0.5557, which indicates that the GMD triplets embeddings learned by HCMGNN are sufficiently discriminative and effective for prediction task.

\begin{figure}[t]
\centering
\includegraphics[width=1\columnwidth]{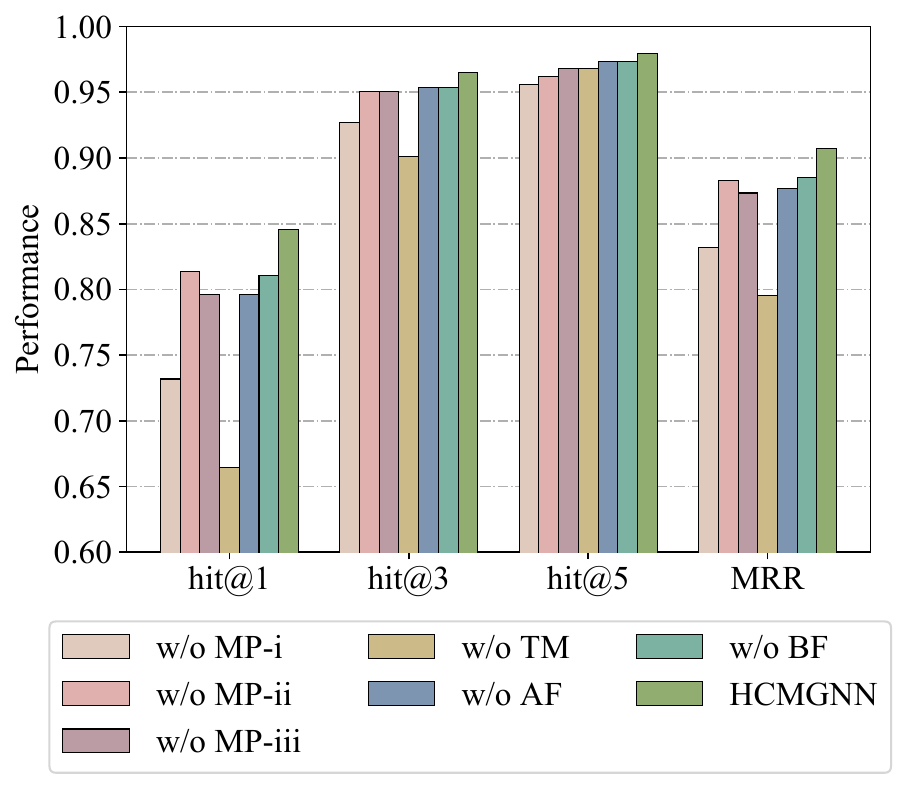}
\caption{Performances of HCMGNN and variants in ablation study.} 
\label{fig2}
\end{figure}

\subsection{Ablation Study} 
To validate the necessity of each component in our model, we further compared several different variants of HCMGNN in independent test experiment.
\begin{itemize}
    \item \textbf{HCMGNN without the intra-subgraph sharing message passing (w/o MP)} \textbf{(w/o MP-\romannumeral1)} updates node embeddings by aggregating metapath-induced neighbors information instead of triple-wise semantic information represented by metapath instances.
    \textbf{(w/o MP-\romannumeral2)} updates node embeddings by aggregating the causal semantic information of each metapath instance only to its head/tail node, instead of sharing information to each node of the instance. 
    \textbf{(w/o MP-\romannumeral3)} defines non-causal metapaths (e.g., \textit{G-M-D-M-G}) with symmetric structure to embody triple-wise information interaction, and updates node embeddings by aggregating non-causal semantics to the head/tail node of each instance.
    \item \textbf{HCMGNN without defined triple-wise interactive metapaths (w/o TM)} defines metapaths of length 2 containing two node types that do not have triple-wise information interactions (e.g., \textit{G-M}). 
    \item \textbf{HCMGNN without inter-subgraph attentive fusion (w/o AF)} adopts the average strategy instead of the attention mechanism to achieve node aggregation between subgraphs.
    \item \textbf{HCMGNN without biological features of nodes (w/o BF)} employs one-hot encoding to replace biological attributes as initial features of nodes.
\end{itemize}

As shown in Figure 3, the performances of all variants of HCMGNN significantly decline, which indicates that these components all contribute to the GMD associations prediction. Specifically, we have the following observations: (1) Compared with HCMGNN (w/o MP-\romannumeral2) and HCMGNN (w/o MP-\romannumeral3), HCMGNN (w/o MP-\romannumeral1) achieves a large performance drop, which shows that only aggregating metapath-induced neighbors information is insufficient for fully capture higher-order relations in heterogeneous graph. Next, the differents result between HCMGNN (w/o MP-\romannumeral2) and HCMGNN (w/o MP-\romannumeral3) suggests that causal metapaths are more helpful in representing triple-wise relations from multiple views than non-causal metapaths with symmetric structures. In addition, HCMGNN outperforms HCMGNN (w/o MP-\romannumeral2) and HCMGNN (w/o MP-\romannumeral3), which indicates that our designed intra-subgraph sharing message passing module is effective for the GMD association prediction. The triple-wise semantic information represented by metapaths cannot simply be aggregated only to the head/tail nodes connecting metapath instances. (2) HCMGNN (w/o TM) has the worst predictive effect across almost all evaluation metrics, demonstrating that metapaths containing only two types of entities are difficult to reflect the entangled interaction of the three entities, which further reveals the capability of our defined causal metapaths to mine high-order relations from heterogeneous graph. (3) HCMGNN (w/o AF) shows performance decline on all evaluation metrics, indicating that different subgraphs divided by causal metapath contribute differently to the model predictions, and inter-subgraph attentive fusion helps to learn multi-view nodes representation comprehensively. (4) HCMGNN (w/o BF) also exhibits a drop in all metrics, suggesting that biological features of genes, microbes, and diseases provide useful domain knowledge for GMD association prediction. 

\subsection{Why Do Causal Metapaths Improve Predictive Performance?}

In this section, we further explore why causal metapaths contribute to the improvement of triple-wise associations predictive performance. Given that metapaths are able to capture high-order relations in heterogeneous graph, we speculate that triple-wise semantic information mined by causal metapaths may have a better impact on low-degree nodes with fewer neighbors (called sparse nodes), which helps to alleviate association sparsity issue caused by sparse nodes. Thus, we calculate the degrees of nodes in heterogeneous graph, and average the degrees for the nodes contained in each triplet to obtain the average node degree (N) for each triplet in independent test set. Next, we divide the test sets into twelve groups by counting triplet numbers under different average node degree interval and evaluate the performances of HCMGNN and HCMGNN without the intra-subgraph sharing message passing (w/o MP) by Hit@1 in the same group. 

As shown in Figure 4, We can observe a consistent performance decline across all models as the average node degree of triplets decreased. This trend indicates that most models exhibit limited capability in aggregating information for sparse nodes. In addition, we also have the following observations: (1) Compared with HCMGNN (w/o MP-\romannumeral1), the predictive performance of HCMGNN is significantly improved in all degree distributions, suggesting that aggregating triple-wise semantic information from multiple causal views is more conducive to learning sparse nodes representations than metapath-induced neighbors. (2) Compared with HCMGNN (w/o MP-\romannumeral2) and HCMGNN (w/o MP-\romannumeral3), the performance gap between HCMGNN and them has notably widened with the decrease of the degree value (see the red area change in Figure 4), which indicates that HCMGNN has superior predictive performance in low-degree triplets, suggesting that causal semantic sharing mechanism enhances the aggregation of information for sparse nodes, effectively alleviating the association sparsity issue in triple-wise association prediction (more details and additional experiments can be found in Appendix Section 6). Based on this result, we infer that causal metapaths have the potential to augment the original semantics and structure of heterogeneous graph, thereby improving the overall predictive performance of triple-wise associations by enhancing the predictive capability for sparse associations.

\begin{figure}[t]
\centering
\includegraphics[width=1\columnwidth]{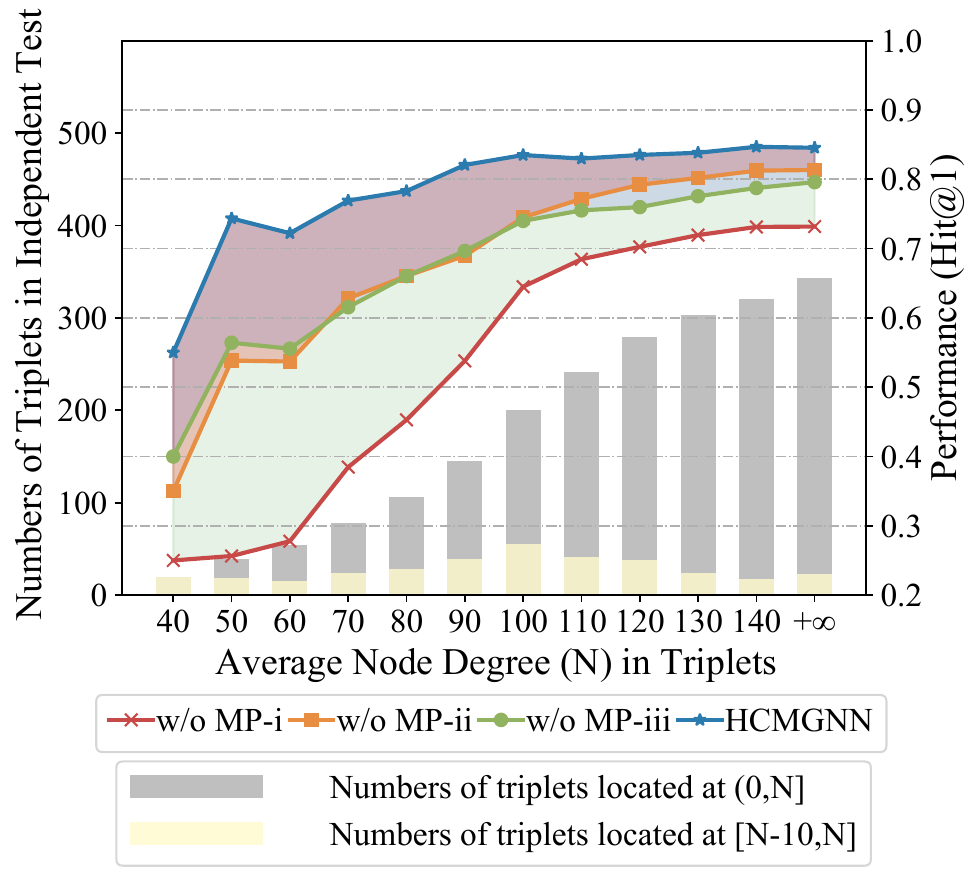} 
\caption{Performances of HCMGNN and w/o MP in test triplets with different average node degree (N) distributions. Each bar shows the result on the triplets in the average node
degree interval (0, N].}
\label{fig3}
\end{figure}

\section{Conclusion}
In this paper, we propose a heterogeneous causal metapath graph neural network for GMD association prediction, named HCMGNN. HCMGNN constructs a heterogeneous graph consisting of genes, microbes, and diseases through their pairwise associations, and then extracts multiple subgraphs according to the patterns of causal metapaths for data augmentation. Next, intra-subgraph message passing with causal semantic sharing mechanism and inter-subgraph attentive fusion are implemented to learn multi-view node representations. Experimental results show that the superior performances of HCMGNN over baselines, and causal metapaths can indeed augment the inherent semantics and structure of heterogeneous graph to alleviate association sparsity issue.

\section*{Acknowledgments}
This work was supported by the National Natural Science Foundation of China (62072206, 62372204, 61772381, 62102158, 32300545); National Key R\&D Program of China (2022YFA1304104); Fundamental Research Funds for the Central Universities (2662021JC008, 2662022JC004); Huazhong Agricultural University Scientific \& Technological Self-innovation Foundation. The funders have no role in study design, data collection, data analysis, data interpretation, or writing of the manuscript.

\bibliographystyle{named}

\end{document}